\documentclass[11pt]{article}
\usepackage{graphicx}
\usepackage[a4paper, top=2.5cm, bottom=2.5cm, left=2cm, right=2cm]{geometry}
\usepackage[english]{babel}
\usepackage{enumitem}
\setlist[itemize]{label=-}
\usepackage{amsmath}
\usepackage{amssymb}
\usepackage{amsthm}
\usepackage{listings}
\usepackage{xcolor}
\usepackage{hyperref}
\usepackage{booktabs}
\usepackage{fancyhdr}

\newtheorem{definition}{Definition}[section]
\newtheorem{remark}{Remark}[section]

\hypersetup{
    colorlinks=true,
    linkcolor=blue,
    filecolor=magenta,
    urlcolor=cyan,
    pdftitle={The Auton Agentic AI Framework},
}

\definecolor{codegray}{rgb}{0.5,0.5,0.5}
\definecolor{codepurple}{rgb}{0.58,0,0.82}
\definecolor{backcolour}{rgb}{0.95,0.95,0.92}

\lstdefinestyle{mystyle}{
    backgroundcolor=\color{backcolour},
    commentstyle=\color{codegray},
    keywordstyle=\color{magenta},
    numberstyle=\tiny\color{codegray},
    stringstyle=\color{codepurple},
    basicstyle=\ttfamily\footnotesize,
    breakatwhitespace=false,
    breaklines=true,
    captionpos=b,
    keepspaces=true,
    numbers=left,
    numbersep=5pt,
    showspaces=false,
    showstringspaces=false,
    showtabs=false,
    tabsize=2
}
\title{\textbf{The Auton Agentic AI Framework} \\ \Large A Declarative Architecture for Specification, Governance, and Runtime Execution of Autonomous Agent Systems}
\author{Sheng Cao \and Zhao Chang \and Chang Li \and Hannan Li \and Liyao Fu \and Ji Tang \\[0.5em]
\normalsize\texttt{\{rcao, zchang, chang.li, hli5, lfu, jtang\}@snapchat.com}}
\date{}

\begin{document}

\maketitle

\begin{abstract}
\noindent The field of Artificial Intelligence is undergoing a transition from \textbf{Generative AI}---probabilistic generation of text and images---to \textbf{Agentic AI}, in which autonomous systems execute actions within external environments on behalf of users. This transition exposes a fundamental architectural mismatch: Large Language Models (LLMs) produce stochastic, unstructured outputs, whereas the backend infrastructure they must control---databases, APIs, cloud services---requires deterministic, schema-conformant inputs. The present paper describes the Auton Agentic AI Framework, a principled architecture for standardizing the creation, execution, and governance of autonomous agent systems. The framework is organized around a strict separation between the \textbf{Cognitive Blueprint}, a declarative, language-agnostic specification of agent identity and capabilities, and the \textbf{Runtime Engine}, the platform-specific execution substrate that instantiates and runs the agent. This separation enables cross-language portability, formal auditability, and modular tool integration via the Model Context Protocol (MCP). The paper formalizes the agent execution model as an augmented Partially Observable Markov Decision Process (POMDP) with a latent reasoning space, introduces a hierarchical memory consolidation architecture inspired by biological episodic memory systems, defines a constraint manifold formalism for safety enforcement via policy projection rather than post-hoc filtering, presents a three-level self-evolution framework spanning in-context adaptation through reinforcement learning, and describes runtime optimizations---including parallel graph execution, speculative inference, and dynamic context pruning---that reduce end-to-end latency for multi-step agent workflows.
\end{abstract}

\section{Introduction}

Large Language Models (LLMs) operate as stochastic inference engines: given an input token sequence, they produce a probability distribution over the next token and sample from it autoregressively~\cite{iti_agentic}. In their default configuration, LLMs lack persistent memory across sessions, provide no deterministic execution guarantees, and impose no structural constraints on multi-step workflows. While these models exhibit facility with natural language, semantic fluency alone does not ensure syntactic or schema compliance in generated outputs. Enterprise deployment demands systems that produce syntactically valid outputs adhering to safety schemas and business logic~\cite{mulani_medium}.

This mismatch between the probabilistic nature of LLM outputs and the deterministic requirements of downstream systems constitutes what is termed here the \textbf{Integration Paradox}. In practice, developers face a binary choice between rigid, hard-coded scripts that do not adapt to novel inputs and opaque agent frameworks whose internal logic resists inspection, testing, and maintenance. No widely adopted standard exists for the formal definition of an agent as a reusable, portable, and auditable unit of autonomous behavior.

The \textbf{Auton Agentic AI Framework} addresses this gap by providing a principled architecture for standardizing the representation, creation, execution, and governance of autonomous agent systems. The central thesis is that the \textbf{Cognitive Blueprint}---the declarative specification of an agent's identity, capabilities, and constraints---must be cleanly separated from the \textbf{Runtime Engine}---the platform-specific execution logic that loads, hydrates, and runs the agent. This separation, analogous to the infrastructure-as-code paradigm established by systems such as Kubernetes and Terraform, enables agents to be specified as versionable, auditable data artifacts independent of any particular programming language or execution environment.

\subsection{Architectural Pillars}

The framework is organized around four principal architectural pillars, each addressing a distinct challenge in the deployment of autonomous agent systems.

\paragraph{The AgenticFormat Standard.}
AgenticFormat is a language-agnostic, declarative schema that adopts a configuration-over-code philosophy for agent definition. The schema specifies an agent's interface, tool bindings, memory configuration, and safety constraints in a structured format (YAML or JSON). Because the definition is decoupled from any particular runtime, an agent specified in a Python development environment can be executed in a high-performance Java runtime without refactoring the agent specification itself.

\paragraph{Deterministic Governance.}
Safety enforcement in the framework is not delegated to prompt engineering or post-hoc output filtering. Instead, the framework introduces a \textbf{Constraint Manifold}---a formally defined subspace of the action space onto which the agent's policy is projected prior to action emission. Policy constraints are expressed as code-level specifications, ensuring that privilege escalation and unsafe operations are excluded by construction rather than detected after the fact.

\paragraph{Cognitive Persistence.}
LLMs are stateless across sessions: when a session terminates or the context window is exhausted, all session-specific experience is lost. The framework addresses this limitation through a hierarchical memory architecture. A \textbf{Reflector-Driven Consolidation Protocol}, drawing on principles from biological memory consolidation, compresses raw event streams into semantic insights. These consolidated memories persist across sessions, enabling agents to incorporate experience from prior interactions without model retraining.

\paragraph{Agentic Efficiency.}
Latency constrains the utility of autonomous agents in interactive and real-time settings. The framework introduces runtime optimizations centered on \textbf{Cognitive Map-Reduce}: the runtime analyzes dependency graphs within agent execution plans and parallelizes independent reasoning and tool-invocation steps, bounding total execution time by the critical path length rather than the sum of all step latencies.

\medskip

The remainder of this paper is structured as follows. Section~\ref{sec:context} characterizes the Integration Paradox and the fragmentation of the current agent development ecosystem. Section~\ref{sec:agenticformat} presents the AgenticFormat Standard and its design principles. Section~\ref{sec:formalism} develops the formal agent execution model. Section~\ref{sec:memory} describes the cognitive memory architecture. Section~\ref{sec:safety} treats safety and governance via the constraint manifold formalism. Section~\ref{sec:evolution} introduces the three-level self-evolution framework. Section~\ref{sec:efficiency} addresses runtime efficiency optimizations. Section~\ref{sec:impact} discusses strategic impact and the open-source roadmap. Section~\ref{sec:conclusion} concludes.

\section{The Integration Paradox and Ecosystem Fragmentation}
\label{sec:context}

\subsection{The Integration Paradox}

The principal obstacle to enterprise adoption of Agentic AI is not model capability per se, but rather a structural mismatch between the output interfaces of LLMs and the input requirements of backend systems. LLMs are probabilistic generators: they produce unstructured or semi-structured text---drafts, summaries, natural-language descriptions---with no formal guarantees on output format. The infrastructure these models must control---relational databases, RPC protocols, cloud APIs, message queues---is deterministic and schema-bound. A single syntax error, type mismatch, or schema violation in an agent-generated command can cause downstream failure. These systems do not accept ambiguous, malformed, or informally specified input.

Consider an autonomous data analyst implemented atop an LLM. When tasked with querying a database, the model may emit a natural-language description of the intended SQL query, or it may produce a query that is semantically plausible but syntactically invalid for the target SQL dialect. The downstream database engine, however, requires a syntactically valid statement executed against a defined schema, authenticated with valid credentials, and subject to safety constraints such as read-only access or row-level security policies. Unstructured or invalid output is not consumable by the database engine and results in a hard failure.

This gap forces developers to introduce layers of ad hoc glue code: regex-based output parsers, retry logic with backoff, format-specific validation layers, and type-coercion routines---all designed to bridge the divide between stochastic model output and the deterministic input contracts of downstream services. The resulting reliability ceiling limits the applicability of agents in mission-critical workflows, where even infrequent failures may be unacceptable.

\subsection{Ecosystem Fragmentation as Technical Debt}

The integration challenge is compounded by fragmentation of the agent development ecosystem into mutually incompatible implementations. In the absence of a unified standard for agent definition, developers face a binary choice, both branches of which incur substantial technical debt:

\begin{enumerate}
    \item \textbf{Rigid, Hard-Coded Workflows.} Agent logic is directly embedded in application code as imperative control flow. Such workflows are brittle: they do not generalize to edge cases, resist modification, and couple agent behavior to the specifics of a single deployment context.
    \item \textbf{Opaque Agent Frameworks.} Frameworks such as LangChain or AutoGen provide higher-level abstractions but conflate the \textit{definition} of an agent---its identity, capabilities, and constraints---with its \textit{runtime execution logic}~\cite{atla_frameworks}. The agent's specification is inseparable from the framework's internal APIs and execution model.
\end{enumerate}

Coupling agent definition to a specific runtime entails vendor lock-in: an agent defined within a Python-centric framework cannot be ported to a different language or execution environment without substantial reimplementation. An agent prototyped in a Python notebook, for instance, cannot be deployed to a Java microservice handling low-latency ad bidding or infrastructure management without rewriting the agent logic in the target language and adapting it to a different runtime model.

These frameworks also lack a standardized, externally auditable schema for agent specification. Safety policies, prompt templates, and tool-binding logic are typically embedded within imperative execution scripts rather than expressed as versioned, machine-readable, and independently reviewable specifications. This entanglement impedes compliance review and makes it difficult to verify that an agent operates within its intended boundaries.

\subsection{Declarative Definitions as a Resolution}
\label{subsec:k8s}

Resolving both the Integration Paradox and the fragmentation problem requires a paradigm shift from \textbf{imperative code} to \textbf{declarative definitions} for agent specification.

A useful parallel exists in the evolution of cloud infrastructure management. Prior to the widespread adoption of container orchestration systems such as Kubernetes, infrastructure was managed through imperative scripts: shell commands, ad hoc configuration files, and manual provisioning steps. Scaling, consistency, and reproducibility were difficult to achieve. Kubernetes introduced a separation between the \textbf{desired state}---declared in structured YAML manifests---and the \textbf{reconciliation loop}---the controller logic that continuously drives the actual system state toward the declared target.

Agentic systems require an analogous architectural separation. The \textbf{Cognitive Blueprint} of an agent---its tools, memory configuration, safety constraints, input/output contracts---must be specified declaratively and decoupled from the \textbf{Runtime Engine} that loads, hydrates, and executes the agent. When agent specifications are expressed as structured data rather than embedded in imperative code, they become strictly typed, versionable, diffable, and auditable. This separation is the foundational premise of the Auton Agentic AI Framework and the AgenticFormat Standard described in the following section.

\section{The AgenticFormat Standard}
\label{sec:agenticformat}

\subsection{Agent Configuration Fragmentation}

Beyond the integration and ecosystem challenges described in Section~\ref{sec:context}, a further structural problem impedes enterprise adoption: the definition of an ``AI Agent'' is tightly coupled to the runtime framework in which it is implemented, rather than conforming to a shared, framework-independent standard~\cite{open_agent_spec}. A LangChain agent is a Python script organized around LangChain's class hierarchy; an AutoGen agent is a distinct set of Python classes with different abstractions; an agent deployed on a high-frequency trading desk may be implemented in C++ with no relation to either framework.

This fragmentation---termed here \textbf{Agent Configuration Balkanization}---produces three concrete consequences for enterprise adoption:

\begin{enumerate}
    \item \textbf{Vendor Lock-in.} Agent definitions are bound to a specific library's API surface. If the library is deprecated, undergoes breaking API changes, or ceases maintenance, all agents defined against that library become difficult or impossible to maintain without rewriting.
    \item \textbf{Auditability Gaps.} When safety logic, prompt templates, and tool definitions are distributed across imperative code paths, compliance review cannot straightforwardly verify agent boundaries, permitted actions, or data access policies. The agent's effective behavior is an emergent property of the code rather than a declared specification.
    \item \textbf{The Polyglot Barrier.} An agent defined in Python cannot execute within a Java microservice environment---or vice versa---without complete reimplementation, forcing organizations to maintain parallel toolchains and duplicated logic across language boundaries.
\end{enumerate}

\subsection{Configuration Over Code}

The Auton Agentic AI Framework treats agents as \textit{data} rather than \textit{code}. The \textbf{AgenticFormat Standard} is a language-agnostic, declarative schema (expressible in YAML or JSON) that defines what is termed the ``Agentic Class'' and establishes a clean separation between the \textbf{Cognitive Blueprint}---the agent's identity, interface, capabilities, and constraints---and the \textbf{Runtime Engine}---the platform-specific execution substrate.

\begin{itemize}
    \item \textbf{The Blueprint (AgenticFormat):} A static, versionable, machine-readable and human-readable specification of the agent's interface, tool bindings, memory constraints, output contracts, and safety manifold. The blueprint is a data artifact; it contains no executable code.
    \item \textbf{The Runtime (Agentic AI Platform SDK):} A platform-specific SDK---\texttt{agentic-py} for Python, \texttt{agentic-java} for Java---that reads a blueprint file and instantiates the corresponding agent within the target execution environment. The SDK is decoupled from the AgenticFormat Standard: the standard defines \textit{what} an agent is, while the SDK provides the runtime machinery to hydrate and execute that definition. The same blueprint can be consumed by any compliant SDK implementation without modification.
\end{itemize}

The design follows the same architectural pattern as Terraform and Kubernetes: as \textbf{Infrastructure-as-Code} describes the desired state of cloud infrastructure in declarative configuration files, \textbf{Agent-as-Configuration} describes the desired state of an autonomous agent. The runtime's responsibility is to reconcile the actual agent state with the declared specification.

\subsection{Contract-Driven Development}

AgenticFormat addresses the Integration Paradox (Section~\ref{sec:context}) through a discipline of \textbf{Contract-Driven Development}. In a system built on probabilistic language generation, ambiguous or unstructured output is a primary failure mode. An agent that performs an automated code review, for example, must not return free-form natural-language text (e.g., ``The code looks fine to me!''); downstream consumers of the agent's output---CI/CD pipelines, review dashboards, issue trackers---expect structured, typed, schema-conformant data.

In AgenticFormat, every agent's output is bound to a formal schema---expressed, for instance, as a YAML inline schema, a Pydantic model, or a JSON Schema definition---rather than an untyped string. Consider a \texttt{Code Reviewer} agent whose task is to review pull requests for correctness, style, and security issues:

\begin{lstlisting}[caption=AgenticFormat Output Contract Definition (Snippet)]
# AgenticFormat Definition (Snippet)
metadata:
  id: code_reviewer
  name: Code Reviewer
  version: 1.2.0
  authors: ["eng-productivity@org.com"]
  tags: [code-quality, automated]

interface:
  input:
    inline_schema: { type: object, properties: { pr_url: { type: string } } }
  output:
    inline_schema: { type: object, properties: { code_ptr_url: { type: string }, review: { type: string } } }

constraints:
  tighten_only_invariant: true
  budget:
    max_token_usage: 50000

action_space:
  mcp_servers:
    - alias: github
      url: https://mcp-github.com
      allow_tools: [get_pr_diff, post_review_comment]
  local_agents:
    - alias: style_checker
      source: ./style-checker.agf.yaml

execution_policy:
  id: x-runtime.react
  config:
    provider: google
    model: gemini-3-pro-preview
    instructions: "Review the PR for correctness, style, and security issues."
    max_steps: 10
    temperature: 0.3
    tool_choice: auto
\end{lstlisting}

With this contract in place, the Runtime Engine interposes a validation layer at the agent's output boundary. If the underlying LLM emits unstructured text or output that violates the declared schema, the runtime detects the violation prior to any downstream propagation and either applies corrective parsing or triggers a retry cycle. The downstream consumer---e.g., a CI/CD pipeline or a review dashboard---receives only valid, schema-conformant output (e.g., \texttt{\{"code\_ptr\_url": "org/repo/code.py\#L1-L10", "review": "Potential null dereference...", ...\}}). The contract thus transforms what would otherwise be a probabilistic, best-effort output channel into a deterministic, typed interface.

\subsection{Integration with the Model Context Protocol}

Agents require access to external tools; integrating each new API endpoint (Slack, GitHub, Postgres, etc.) typically demands custom glue code, bespoke authentication handling, and ad hoc data marshalling. The AgenticFormat Standard addresses this by adopting the \textbf{Model Context Protocol (MCP)}~\cite{mcp_anthropic, mcp_explainer} as the standard mechanism for tool integration.

The division of responsibilities between the two standards is as follows:

\begin{itemize}
    \item \textbf{MCP} standardizes \textit{how} an agent connects to external services. A tool connector implemented to the MCP specification---e.g., a Google Drive connector or a Slack connector---is usable by any MCP-compatible agent, regardless of the agent's definition framework.
    \item \textbf{AgenticFormat} standardizes \textit{who} uses the tools: the agent's identity, permission scope, allowed tool set, and the specific MCP servers to which the agent is bound.
\end{itemize}

The combination of these two standards supports modular, composable system design. A \texttt{Customer Support Agent} can be defined in AgenticFormat and bound to a \texttt{SalesforceMCP} server for CRM access. Migrating from Salesforce to HubSpot requires only substituting the MCP server binding in the agent's blueprint; the agent's cognitive specification---its reasoning strategy, memory configuration, output contracts, and safety constraints---remains unchanged.

\section{Formal Agent Execution Model}
\label{sec:formalism}

\subsection{The Agent as a Control System}

Anthropomorphic characterizations of autonomous agents---``digital employees,'' ``AI assistants''---obscure the stochastic nature of the underlying inference process and impede rigorous analysis of system behavior. The Auton Agentic AI Framework instead models an agent as a decision-making system operating within a \textbf{Partially Observable Markov Decision Process (POMDP)}~\cite{kaelbling1998pomdp, llm_guided_pomdp}, augmented with a latent reasoning mechanism that decouples internal computation from external action~\cite{microsoft_reasoning}.

A standard reinforcement learning (RL) agent implements a reactive mapping from observations to actions ($O \rightarrow A$). An Agentic System, as formalized here, interposes a \textbf{Latent Reasoning Space} ($\mathcal{Z}$) between observation and action. Within $\mathcal{Z}$, the system can plan, reflect, and verify candidate action sequences \textit{without} altering the external environment state. This architectural choice separates internal deliberation from externally visible side effects and provides a formal basis for the ``think-before-act'' execution discipline described below.

\subsection{The Augmented POMDP Formulation}

\begin{definition}[Agentic System Tuple]
\label{def:agentic-tuple}
An \textbf{Agentic System} is defined by the augmented tuple $\mathcal{T}$:
\begin{equation}
\mathcal{T} = \langle \mathcal{S}, \Omega, \mathcal{A}, \mathcal{Z}, \mathcal{M}, \mathcal{P}, \mathcal{R} \rangle
\end{equation}
where each component is defined as follows.
\end{definition}

\paragraph{Latent World State ($\mathcal{S}$).}
The set of all possible true environment states. The world state $s \in \mathcal{S}$ encompasses the full configuration of external systems---database contents, server load, network state, user session state---and is not directly observable by the agent. The agent must infer a belief distribution over $\mathcal{S}$ from partial observations.

\paragraph{Observation Space ($\Omega$).}
The set of partial views available to the agent at each timestep: API responses, search results, error messages, sensor readings, and other environment signals. Given true state $s_t$, the agent receives observation $o_t \sim O(o_t | s_t)$, where $O$ is the observation function. The agent uses its observation history to maintain a state estimate---a belief distribution over $\mathcal{S}$---that approximates the unobservable true state.

\paragraph{External Action Space ($\mathcal{A}$).}
The set of actions that produce side effects in the external environment: database mutations, API calls, file system operations, message transmissions. Actions in $\mathcal{A}$ are subject to the safety constraints imposed by the Constraint Manifold (Section~\ref{sec:safety}). Each external action $a \in \mathcal{A}$ transitions the environment from state $s$ to a successor state $s'$ according to the transition kernel $\mathcal{P}$.

\paragraph{Latent Reasoning Space ($\mathcal{Z}$).}
The set of internal cognitive operations: planning, reflection, self-verification, hypothesis generation. Actions in $\mathcal{Z}$ consume computational resources (tokens, wall-clock time) but do not alter the external state $\mathcal{S}$. The existence of $\mathcal{Z}$ as a formally distinct component of the tuple enforces the separation between deliberation and action at the architectural level.

\paragraph{Memory Context ($\mathcal{M}$).}
The agent's internal state, comprising the current observation history $\mathcal{H}_t = (o_0, a_0, z_0, \ldots, o_t)$ together with consolidated knowledge retrieved from long-term storage. The memory context $m_t \in \mathcal{M}$ serves as the sufficient statistic upon which the agent conditions its reasoning and action policies at each timestep.

\paragraph{Transition Kernel ($\mathcal{P}$).}
The environment dynamics function $T(s' | s, a)$, specifying the probability of transitioning to state $s'$ given current state $s$ and action $a \in \mathcal{A}$. Internal reasoning actions $z \in \mathcal{Z}$ do not induce state transitions: $T(s' | s, z) = \delta(s' - s)$ for all $z \in \mathcal{Z}$.

\paragraph{Reward Function ($\mathcal{R}$).}
A function $\mathcal{R}: \mathcal{S} \times \mathcal{A} \times \mathcal{Z} \rightarrow \mathbb{R}$ assigning scalar feedback to state-action-reasoning triples. \textbf{Sparse rewards} evaluate outcomes (e.g., binary task success or failure at the terminal state), while \textbf{dense rewards} provide step-level process feedback (e.g., intermediate correctness checks on reasoning steps). The reward function may also penalize inefficient reasoning: excessive token expenditure, repetitive computation, or looping behavior within $\mathcal{Z}$.

\subsection{Factorized Policy Architecture}

In standard RL formulations and in direct LLM-to-action pipelines, the policy $\pi(a|s)$ implements a reactive mapping from observation to action. This reflexive architecture is computationally efficient but error-prone: the agent has no mechanism to evaluate, compare, or verify candidate actions before committing to one.

The Auton Agentic AI Framework replaces the monolithic policy with a \textbf{Factorized Policy Architecture}~\cite{yao2023react} that decomposes agent behavior into two coupled sub-policies, thereby enforcing a \textbf{think-before-act} invariant at the architectural level. At each timestep $t$, the agent's execution proceeds in two stages.

\subsubsection{The Reasoning Policy ($\pi_{\text{reason}}$)}

The agent first samples a reasoning trace $z_t$ from the latent space $\mathcal{Z}$, conditioned on the current memory context $m_t$:
\begin{equation}
z_t \sim \pi_{\text{reason}}(z_t \mid m_t; \theta)
\end{equation}
where $\theta$ parameterizes the reasoning policy. The reasoning trace $z_t$ may take the form of a chain-of-thought decomposition~\cite{wei2022cot}, a planning step, a self-critique, or a verification check against known constraints. Execution of $\pi_{\text{reason}}$ updates the memory context $m_t$ but produces no external side effects. This stage enables \textbf{test-time compute scaling}: the agent can sample multiple candidate reasoning paths, evaluate them against internal criteria, and select a plan before committing to an external action.

\subsubsection{The Action Policy ($\pi_{\text{action}}$)}

Conditioned on both the current memory context $m_t$ and the generated reasoning trace $z_t$, the agent samples an external action $a_t$:
\begin{equation}
a_t \sim \pi_{\text{action}}(a_t \mid m_t, z_t; \phi)
\end{equation}
where $\phi$ parameterizes the action policy. The conditioning on $z_t$ ensures that no external action is taken without a preceding deliberation step. By construction, the action is informed by the reasoning trace, which reduces the incidence of impulsive or poorly considered actions relative to a monolithic policy that maps directly from observation to action.

\begin{remark}
The factorized architecture does not require two separate neural networks; in practice, both $\pi_{\text{reason}}$ and $\pi_{\text{action}}$ may be implemented by the same LLM, with the factorization enforced by the runtime's execution protocol rather than by model architecture.
\end{remark}

\subsection{Objective Function}

The agent maximizes expected discounted return over joint trajectories of latent reasoning traces and external actions. Let $\tau = (o_0, z_0, a_0, r_0, o_1, z_1, a_1, r_1, \ldots)$ denote a trajectory generated under the joint policy $(\pi_{\text{reason}}, \pi_{\text{action}})$. The optimization objective $J(\theta, \phi)$ is:
\begin{equation}
J(\theta, \phi) = \mathbb{E}_{\tau \sim (\pi_{\text{reason}}, \pi_{\text{action}})} \left[ \sum_{t=0}^{T} \gamma^t R(s_t, a_t, z_t) \right]
\end{equation}
where $\gamma \in [0, 1)$ is the discount factor and $T$ is the episode horizon. The reward function $R(s_t, a_t, z_t)$ decomposes into a task-completion component $R(s_t, a_t)$ and a reasoning-efficiency component $R(z_t)$. The latter can penalize inefficient reasoning patterns---repetitive traces, circular logic, or excessive token expenditure---thereby incentivizing concise, goal-directed deliberation.

\section{Cognitive Memory Architecture}
\label{sec:memory}

\subsection{The Statelessness Limitation}

LLMs possess large \textbf{parametric knowledge}---factual and procedural information encoded in model weights during pre-training---but lack persistent \textbf{episodic memory}. When a session terminates or the context window is exhausted, all session-specific information---observations, tool outputs, intermediate reasoning---is discarded. The model has no native mechanism to retain or recall experience from prior interactions.

For autonomous systems that are expected to improve performance over time, this statelessness is a fundamental limitation. Naive approaches to persistence, such as appending raw interaction logs to the context window, do not scale: the resulting context grows without bound, eventually exceeding the model's context window and imposing quadratic attention cost even within the window limit. The Auton Agentic AI Framework addresses this limitation through a hierarchical memory architecture inspired by \textbf{biological memory consolidation}~\cite{princeton_consolidation, memory_replay_bio}: information is transferred from transient, high-fidelity buffers to persistent, compressed storage through a structured consolidation process~\cite{aws_memory}.

\subsection{Hierarchical Memory Structure}

The agent's memory $\mathcal{M}$ is organized into two coupled layers, distinguished by temporal scope, fidelity, and access characteristics.

\subsubsection{Short-Term Memory (Event Stream)}

Short-term memory constitutes the agent's \textbf{working memory}: a high-fidelity, temporally ordered log of the current interaction session.

\begin{itemize}
    \item \textbf{Content:} The event stream records UserEvents (user prompts and inputs), ToolCallEvents (API requests and responses), and SystemLogs (error traces, status messages, intermediate outputs).
    \item \textbf{Characteristics:} The event stream is ephemeral, strictly temporally ordered, and bounded by the model's context window (e.g., 128k tokens for current-generation models). When the context window is exhausted, the oldest entries must be evicted or compressed.
\end{itemize}

\subsubsection{Long-Term Memory (Knowledge Base)}

Long-term memory provides persistent storage that survives across sessions. It is subdivided into three functionally distinct stores:

\begin{itemize}
    \item \textbf{Semantic Memory:} General facts about the world and about the user's environment and preferences (e.g., ``The user requires all SQL queries to target read-only replicas''). Semantic memories are stable, context-independent facts that inform the agent's default behavior.
    \item \textbf{Episodic Memory:} Compressed records of past interaction episodes, indexed by embedding vectors for similarity-based retrieval~\cite{semantic_episodic_biorxiv}. Each episodic memory encodes a specific experience and its outcome (e.g., ``In task \#402, applying \texttt{pandas} to a 5\,GB CSV file triggered an out-of-memory error; using \texttt{polars} with lazy evaluation resolved the issue'').
    \item \textbf{Procedural Memory:} Stored action sequences---reusable plans or templates---that encode effective solutions to recurring problem types. Procedural memories represent compiled expertise: multi-step workflows that have been validated through prior execution.
\end{itemize}

\subsection{Consolidation Protocol}

Information is transferred from short-term to long-term memory via a \textbf{Reflector-Driven Consolidation Protocol}. The protocol draws on the concept of \textbf{hippocampal replay} from neuroscience~\cite{memory_replay_bio, princeton_consolidation}: the offline replaying and compression of recent experience into stable long-term representations~\cite{memo_embodied}.

When a session terminates or the context window approaches capacity, the \textbf{Reflector Agent}---a dedicated background process---executes the following consolidation procedure:

\begin{enumerate}
    \item \textbf{Event Segmentation.} The Reflector partitions the raw event stream into coherent logical episodes, where each episode corresponds to a self-contained sub-task or interaction phase (e.g., ``Attempting to query the production database,'' ``Handling an authentication timeout error'').
    \item \textbf{Insight Extraction.} For each identified episode, the Reflector extracts salient \textbf{insights}---observations, outcomes, and causal relationships judged to have high utility for future tasks. Low-utility content (e.g., conversational greetings, boilerplate acknowledgments, redundant intermediate outputs) is discarded. The extraction criterion is estimated future informativeness: content is retained in proportion to its expected contribution to future task performance.
    \item \textbf{Vectorization and Storage.} Extracted insights are embedded into a vector representation and stored in either a long-term knowledge graph or a vector store configured for semantic similarity retrieval. The embedding allows efficient retrieval based on semantic proximity to the agent's current context at query time.
    \item \textbf{Context Compression.} The raw event stream in the active context window is replaced by a compressed summary that preserves the essential information content of the discarded entries. This operation frees token budget within the context window while retaining sufficient information for long-horizon coherence in the agent's reasoning.
\end{enumerate}

\subsection{Formal Characterization}

The consolidation process can be characterized as an information-theoretic optimization: the objective is to minimize the information loss incurred by compressing the full interaction history $h$ into a compact memory representation $m$, subject to a storage budget constraint. Formally, the KL divergence between the distribution induced by the full history $P(h)$ and the distribution induced by the compressed memory $P(m)$ should be minimized.

Equivalently, the consolidation objective can be stated as maximizing the \textbf{mutual information} between the compressed memory $m$ and future tasks:
\begin{equation}
\max_m \; I(m; \text{future\_task})
\end{equation}
subject to a constraint on the size of $m$. Under this formulation, the memory system is optimized to retain precisely the information that maximizes the agent's expected success probability on future, unseen tasks. Content that is predictable from the agent's parametric knowledge or that has low variance across tasks contributes little mutual information and is preferentially compressed or discarded.

\section{Safety and Governance}
\label{sec:safety}

\subsection{Limitations of Post-hoc Filtering}

Many existing approaches to AI safety in agentic systems rely on post-hoc output filtering: the agent generates an action, a separate validation module checks the action against a set of rules (e.g., regular expressions, keyword blocklists, or classifier-based detectors), and the action is blocked if it violates policy. This architecture is inherently fragile. The validation layer operates on the output of an unconstrained generation process; it must anticipate and enumerate all possible policy violations, a task whose difficulty grows combinatorially with the complexity of the action space. Safety is imposed as an afterthought rather than as a structural property of the system.

The Auton Agentic AI Framework replaces post-hoc filtering with \textbf{policy projection}~\cite{safe_rl_manifold}. Rather than generating actions in an unconstrained space and then filtering, the agent's policy is projected onto a formally defined safe subspace---the \textbf{Constraint Manifold}---\textit{prior to} action emission~\cite{formal_methods_verification}. Under this architecture, unsafe actions are not generated and subsequently blocked; they are assigned zero probability during generation.

\subsection{The Constraint Manifold}

Let the full external action space be $\mathcal{A}$ (e.g., the set of all syntactically valid SQL statements). The \textbf{safe sub-manifold} $\mathcal{C} \subset \mathcal{A}$ is defined as the subset of actions that satisfy all applicable enterprise invariants---for example, read-only access restrictions, row-level security policies, data residency requirements, or prohibitions on personally identifiable information (PII) egress~\cite{deny_monotone_access}.

\begin{definition}[Constraint Manifold]
\label{def:constraint-manifold}
Given an action space $\mathcal{A}$ and a set of safety predicates $\{c_1, c_2, \ldots, c_k\}$ where each $c_i: \mathcal{A} \rightarrow \{0, 1\}$, the constraint manifold $\mathcal{C}$ is the intersection:
\begin{equation}
\mathcal{C} = \{ a \in \mathcal{A} \mid c_i(a) = 1 \;\; \forall \, i \in \{1, \ldots, k\} \}
\end{equation}
\end{definition}

Rather than sampling from the unconstrained (``raw'') policy $\pi_{\text{raw}}$ and then validating the sample, the framework projects $\pi_{\text{raw}}$ onto $\mathcal{C}$ to obtain a safe policy $\pi_{\text{safe}}$. The projection is defined by re-normalizing $\pi_{\text{raw}}$ over $\mathcal{C}$:
\begin{equation}
\pi_{\text{safe}}(a|s) = \frac{\pi_{\text{raw}}(a|s) \cdot \mathbb{I}[a \in \mathcal{C}]}{\int_{\mathcal{A}} \pi_{\text{raw}}(x|s) \cdot \mathbb{I}[x \in \mathcal{C}] \, dx}
\end{equation}
where $\mathbb{I}[\cdot]$ is the indicator function. The denominator is the total probability mass that $\pi_{\text{raw}}$ assigns to safe actions; re-normalization ensures that $\pi_{\text{safe}}$ is a valid probability distribution concentrated entirely on $\mathcal{C}$.

\begin{remark}
The projection preserves the relative likelihood ordering among safe actions: if $\pi_{\text{raw}}$ assigns higher probability to action $a$ than to action $a'$ within $\mathcal{C}$, then $\pi_{\text{safe}}$ preserves this ordering. The projection removes probability mass from unsafe actions without distorting preferences among safe alternatives.
\end{remark}

\paragraph{Implementation.}
In the AgenticFormat runtime, the constraint manifold $\mathcal{C}$ is enforced through a \textbf{masking function} $M(s, a)$ applied during autoregressive token generation. The masking function sets the logits of tokens that would lead to unsafe action completions to $-\infty$, ensuring that after the softmax operation, unsafe token sequences receive zero probability. This constrained decoding mechanism operates at the token level during generation, not as a post-hoc check on completed outputs, and guarantees that all emitted actions lie within $\mathcal{C}$ by construction.

\subsection{Economic Constraints via KKT Conditions}

Safety governance in the framework extends beyond action-space restrictions to encompass \textbf{computational economics}: unbounded reasoning loops or unconstrained token consumption impose both direct financial cost and indirect risk through increased latency and resource contention.

The agent operates under a token budget $B$, which constitutes a hard constraint on total token expenditure per task or session. The optimization objective is to maximize task reward $J(\theta)$ subject to the constraint that total token cost $C(\theta)$ does not exceed $B$.

This is a constrained optimization problem. Introducing a Lagrange multiplier $\lambda \geq 0$ yields the Lagrangian $\mathcal{L}$:
\begin{equation}
\mathcal{L}(\theta, \lambda) = J(\theta) - \lambda \big( C(\theta) - B \big)
\end{equation}

The optimal policy $\theta^*$ and the associated multiplier $\lambda^*$ satisfy the \textbf{Karush-Kuhn-Tucker (KKT) conditions}:

\begin{enumerate}
    \item \textbf{Stationarity:} $\nabla_\theta J(\theta^*) = \lambda^* \nabla_\theta C(\theta^*)$. At the optimum, the marginal increase in task reward per unit of policy change equals $\lambda^*$ times the marginal increase in token cost. The multiplier $\lambda^*$ thus quantifies the marginal value (shadow price) of an additional token.
    \item \textbf{Primal Feasibility:} $C(\theta^*) \le B$. The optimal policy does not exceed the token budget.
    \item \textbf{Dual Feasibility:} $\lambda^* \ge 0$. The shadow price of tokens is non-negative.
    \item \textbf{Complementary Slackness:} $\lambda^* (C(\theta^*) - B) = 0$. If the budget is not fully consumed ($C(\theta^*) < B$), then $\lambda^* = 0$ and the budget constraint is inactive---additional tokens have zero marginal cost. If the budget is exactly consumed ($C(\theta^*) = B$), then $\lambda^* > 0$ and each additional token consumed must be justified by a commensurate increase in task reward.
\end{enumerate}

\paragraph{Implementation.}
The runtime's \textbf{Budget Controller} operationalizes this formalism. When token consumption is low relative to task progress---i.e., the budget constraint is slack---the effective $\lambda$ is near zero, and the reasoning policy $\pi_{\text{reason}}$ is permitted to generate expansive chains of thought, explore multiple reasoning paths, and perform thorough verification. As token consumption $C(\theta)$ approaches the budget $B$, the effective $\lambda$ increases, and the reasoning policy is biased toward brevity, directness, and immediate action. This adaptive mechanism balances reasoning depth against resource consumption without requiring manual tuning of reasoning length.

\section{Self-Evolving Agents and End-to-End Optimization}
\label{sec:evolution}

\subsection{Limitations of Static Agent Configurations}

Current deployment practice relies predominantly on ``frozen'' agents: the underlying LLM is pre-trained offline, and agent behavior is determined by static prompts, fixed tool configurations, and hand-crafted decision logic. When such a statically configured agent fails on a task, a human operator must manually diagnose the failure, adjust the prompt or tool configuration, and redeploy. This human-in-the-loop optimization process is labor-intensive and does not scale to large agent deployments or rapidly changing environments.

The Auton Agentic AI Framework treats deployed agents as \textbf{learnable control systems} rather than static configurations~\cite{survey_self_evolving, sage_agent}. \textbf{End-to-End (E2E) Agentic Training} is proposed: the agent updates its effective policy through structured interaction with its environment, closing the feedback loop between reasoning, action, and observed outcomes~\cite{agentic_rl_demystified}.

\subsection{Training Loop Formulation}

Agent training differs from the training paradigm used for conversational chatbots (e.g., Reinforcement Learning from Human Feedback, RLHF). In chatbot training, the objective is \textbf{preference alignment}: generating outputs that humans judge as plausible, helpful, or stylistically appropriate. In agentic training, the objective is \textbf{utility}: successful completion of tasks with measurable, often binary, outcomes.

The E2E agentic training loop comprises three stages:

\begin{itemize}
    \item \textbf{Trajectory Generation.} The agent attempts a task $\mathcal{T}$ within its environment, generating a complete trajectory $\tau = (o_0, z_0, a_0, o_1, z_1, a_1, \ldots, s_T)$ that records the full sequence of observations, reasoning traces, and actions through task completion or failure.
    \item \textbf{Outcome Evaluation (Sparse Reward).} A verifier module evaluates the terminal state $s_T$ of the trajectory and assigns a reward signal $R_{\text{outcome}} \in \{0, 1\}$ (or a scalar value in the continuous case). The verifier may be a deterministic unit test, a compiler, a database constraint checker, or a stronger ``Teacher'' LLM that grades the outcome against a rubric.
    \item \textbf{Process Supervision (Dense Reward).} For credit assignment in long-horizon tasks---where the sparse terminal reward provides insufficient signal to identify which intermediate steps contributed to success or failure---\textbf{Process Reward Models (PRMs)}~\cite{lightman2023prm} or a \textbf{Reflector Agent} evaluate individual reasoning steps $z_t$, producing a dense reward signal $R_{\text{process}}(z_t)$ that enables fine-grained attribution~\cite{neurips_multi_turn}.
\end{itemize}

The composite reward function $R(\tau)$ over a full trajectory is a weighted combination of outcome and process rewards:
\begin{equation}
R(\tau) = R_{\text{outcome}}(\tau) + \lambda \sum_{t=0}^{T} R_{\text{process}}(z_t)
\end{equation}
where $\lambda$ controls the relative contribution of process supervision to the total training signal.

\subsection{Three Levels of Agentic Evolution}

Agentic self-improvement is organized into three levels, ordered by increasing permanence of the adaptation and increasing computational cost.

\subsubsection{Level 1: In-Context Evolution}

\paragraph{Mechanism.}
At this level, the agent adapts without any modification to its model weights~\cite{shinn2023reflexion, langchain_reflection}. When the agent fails a task (generating a failed trajectory $\tau_{\text{fail}}$), the Reflector Agent analyzes the execution trace, identifies the root cause of the failure (e.g., ``incorrect date format used in API call''), and generates a textual ``Lesson'' $L$ that encodes the corrective insight.

\paragraph{Storage and Retrieval.}
The lesson $L$ is stored in Long-Term Memory (Section~\ref{sec:memory}). When the agent subsequently encounters a task with similar characteristics---as determined by embedding-based similarity search---it retrieves $L$ and prepends it to its active context window, effectively conditioning its behavior on prior experience.

\paragraph{Formal Characterization.}
The effective policy under in-context evolution is conditioned on retrieved lessons:
\begin{equation}
\pi_{\text{effective}}(a|s) = \pi_{\theta}(a \mid s, m, L)
\end{equation}
Behavior is modified through the memory mechanism rather than through gradient-based parameter updates. This is analogous to one-shot or few-shot in-context learning, where the model's behavior shifts in response to contextual examples without weight modification.

\subsubsection{Level 2: Self-Taught Reasoning (STaR)}

\paragraph{Mechanism.}
Level 2 internalizes successful reasoning patterns into the model weights via Supervised Fine-Tuning (SFT). The procedure, based on the Self-Taught Reasoner (STaR) framework~\cite{star_reasoner}, operates as follows:

\begin{enumerate}
    \item The agent generates multiple reasoning trajectories $\{ \tau_1, \tau_2, \ldots, \tau_k \}$ for a dataset of training tasks.
    \item \textbf{Filtering:} A ground-truth oracle (e.g., unit tests, formal verifiers) evaluates each trajectory. Only trajectories that produce correct outcomes are retained~\cite{cleaner_trajectory}. This filtering step creates a ``self-purified'' dataset of (Problem, Rationale, Solution) triplets generated by the agent itself.
    \item \textbf{Fine-Tuning:} The model is fine-tuned on its own successful reasoning traces, minimizing the negative log-likelihood:
    \begin{equation}
    \mathcal{L}_{\text{SFT}} = - \sum_{(s, z, a) \in \tau_{\text{success}}} \log \pi_{\theta}(z, a \mid s)
    \end{equation}
\end{enumerate}

This procedure converts slow, deliberative reasoning processes---those requiring extended search, backtracking, or multiple attempts---into fast, single-pass heuristics encoded in the model weights. Over successive iterations, complex multi-step reasoning patterns that initially required exhaustive exploration become directly accessible as learned routines.

\subsubsection{Level 3: Agentic Reinforcement Learning}

\paragraph{Mechanism.}
Level 3 employs on-policy Reinforcement Learning---specifically \textbf{Group Relative Policy Optimization (GRPO)}~\cite{shao2024grpo, guo2025deepseekr1} or \textbf{Proximal Policy Optimization (PPO)}~\cite{schulman2017ppo} adapted for multi-turn POMDPs~\cite{k_level_policy}---to discover execution strategies that may not appear in any existing training data. Whereas SFT at Level 2 distills and compresses known-good trajectories, RL at Level 3 enables exploration beyond the support of the training distribution. The agent can discover novel strategies (e.g., checking a cache before issuing a database query, or parallelizing independent API calls) that maximize expected reward.

\paragraph{Formal Characterization.}
The policy gradient update for the reasoning policy $\pi_{\theta}$ maximizes the expected advantage $\hat{A}_t$ of each reasoning step:
\begin{equation}
\nabla_\theta J \approx \frac{1}{N} \sum_{i=1}^{N} \sum_{t=1}^{T} \nabla_\theta \log \pi_\theta(z_t^{(i)}|h_t^{(i)}) \hat{A}_t^{(i)}
\end{equation}
where $N$ is the number of sampled trajectories, $h_t^{(i)}$ is the history at timestep $t$ in trajectory $i$, and $\hat{A}_t^{(i)}$ is the estimated advantage at that timestep. The resulting agent can discover execution strategies that are more efficient than any trajectory present in the supervised training set.

\subsection{Compounding Improvement and Data Accumulation}

The three-level framework produces a compounding improvement cycle through progressive data accumulation.

\begin{itemize}
    \item \textbf{Stage 1:} The agent operates at Level 1, using the Reflector to accumulate a proprietary database of ``Lessons Learned'' and ``Edge Cases'' specific to the enterprise's data, systems, and workflows.
    \item \textbf{Stage 2:} This accumulated database serves as training data for Level 2 (STaR), producing a fine-tuned specialist model that outperforms generic frontier models on company-specific tasks by virtue of its internalized domain knowledge.
    \item \textbf{Stage 3:} Continuous Level 3 (RL) optimization ensures that the agent adapts to evolving business logic, API schema changes, and shifting data distributions, reducing ongoing maintenance costs.
\end{itemize}

Each stage generates data that feeds subsequent stages, creating a self-reinforcing loop in which operational experience translates into progressively more capable and specialized agent behavior.

\section{Inference Efficiency}
\label{sec:efficiency}

\subsection{The Latency Problem}

The preceding sections addressed the \textit{quality} and \textit{safety} of agentic reasoning; this section addresses \textit{deployment viability}. In interactive and real-time applications, latency---not model capability---is frequently the binding constraint on agent utility.

In a synchronous execution loop, each of $k$ sequential steps (reasoning $\rightarrow$ tool invocation $\rightarrow$ observation $\rightarrow$ reasoning) contributes additively to total end-to-end latency:
\begin{equation}
L_{\text{total}} = \sum_{i=1}^{k} \big( L_{\text{inference},i} + L_{\text{network},i} \big)
\end{equation}
where $L_{\text{inference},i}$ is the model inference latency at step $i$ and $L_{\text{network},i}$ is the network round-trip time for the tool call at step $i$. For workflows involving slow external APIs, large database queries, or deep reasoning chains, total latency can reach minutes---far beyond acceptable thresholds for real-time or near-real-time applications.

The Auton Agentic AI Framework replaces the sequential execution chain with \textbf{asynchronous graph execution}, reducing end-to-end latency by exploiting parallelism and speculation within the agent's execution plan.

\subsection{Cognitive Map-Reduce}

Many agent frameworks execute plan steps in strict sequential order, even when steps have no data dependencies on one another. The Auton Agentic Runtime incorporates a \textbf{Dependency Analyzer} that interprets the agent's execution plan as a \textbf{Directed Acyclic Graph (DAG)} rather than a linear sequence.

When the agent produces a multi-step plan---e.g., check AAPL stock price, check MSFT sentiment score, then compare the two---the runtime's dependency analyzer identifies that the AAPL and MSFT lookups are independent and can execute concurrently (the ``Cognitive Map'' phase). Results from independent subtasks are then aggregated and passed to dependent steps (the ``Cognitive Reduce'' phase).

Under this execution model, total wall-clock time is bounded by the \textbf{critical path}---the longest chain of sequentially dependent steps---rather than by the sum of all step latencies:
\begin{equation}
L_{\text{total}} = \max_{\text{path} \in \text{DAG}} \left( \sum_{\text{node} \in \text{path}} L_{\text{node}} \right)
\end{equation}
For wide, shallow task graphs---e.g., researching multiple entities in parallel, querying multiple data sources concurrently---the critical path length may be substantially shorter than the total sum of all node latencies, yielding proportional reductions in wall-clock execution time.

\subsection{Speculative Execution}

To mitigate the latency imposed by slow external tools (e.g., long-running database queries, API calls with high network latency), the runtime employs \textbf{speculative execution}, analogous to branch prediction in processor microarchitecture.

In a standard synchronous execution loop, the model idles while awaiting tool output. In the \textbf{Speculative Runtime}, the model does not wait; instead, it proceeds optimistically:

\begin{itemize}
    \item \textbf{Prediction.} While the tool call executes asynchronously, the model generates a prediction of the likely tool output---e.g., a success indication, a schema-compliant placeholder result, or the most probable output given the query.
    \item \textbf{Lookahead.} The model computes the next reasoning step conditioned on the predicted tool output, effectively executing ahead of the actual result.
    \item \textbf{Commit or Rollback.} When the actual tool output arrives, the runtime compares it to the prediction. If the actual output matches the prediction (or is semantically equivalent within a defined tolerance), the precomputed lookahead tokens are committed and execution continues without interruption. If the actual output diverges from the prediction, the lookahead branch is discarded and the model regenerates from the actual output.
\end{itemize}

This \textbf{optimistic concurrency} strategy hides a portion of external tool latency behind the model's inference computation. When tool outputs are predictable---as is often the case for well-structured API calls---the speculative branch is committed with high probability, and the effective latency reduction is substantial.

\subsection{Dynamic Context Pruning}

Self-attention cost scales quadratically, $O(N^2)$, in the context length $N$. Retaining every log entry, error trace, intermediate reasoning step, and tool output in the active context saturates the context window and imposes increasing cost and latency as the session progresses.

The framework employs a \textbf{dynamic KV-cache eviction policy} driven by attention-score analysis to maintain a bounded active context size:

\begin{itemize}
    \item \textbf{Semantic dead-weight detection.} The runtime inspects attention patterns in the final $k$ transformer layers. Tokens that consistently receive low attention scores across recent inference steps---e.g., boilerplate headers in prior tool outputs, or greetings from earlier in the conversation---are identified as candidates for eviction.
    \item \textbf{Attention-guided pruning.} Low-attention tokens are evicted from the KV cache based on their aggregate attention scores, rather than by a simple first-in-first-out (FIFO) policy. This ensures that semantically relevant but temporally distant tokens are retained, while recent but uninformative tokens are discarded.
    \item \textbf{Cold storage.} Content that is assessed as potentially relevant for future retrieval but is not actively attended to in the current reasoning context is migrated to a vector store. The active context window thus remains bounded, while previously evicted information remains retrievable via embedding-based similarity search if needed.
\end{itemize}

Under this policy, inference cost remains bounded as session length grows, avoiding the unbounded quadratic scaling that would otherwise render long-running agent sessions prohibitively expensive.

\section{Strategic Impact and Open-Source Roadmap}
\label{sec:impact}

\subsection{A Unified Governance Layer}

The \textbf{AgenticFormat Standard} and the \textbf{Agentic AI Platform SDK} serve distinct and deliberately decoupled roles within the framework. The AgenticFormat Standard is a broad definition protocol: it specifies the schema, constraints, and contracts that constitute an agent's identity, independent of any particular runtime or programming language. The Agentic AI Platform SDK, by contrast, is the execution-side complement: it reads an AgenticFormat blueprint and instantiates a running agent within a specific target environment. The standard defines \textit{what} an agent is; the SDK determines \textit{how} that definition is brought to life.

This decoupling ensures that the definition layer and the execution layer can evolve independently. The AgenticFormat Standard can be adopted by third-party runtimes, proprietary platforms, or alternative SDK implementations without requiring changes to the standard itself. Conversely, the SDK can incorporate runtime optimizations, new model integrations, or platform-specific capabilities without altering the agent specifications it consumes.

The current ecosystem consists of incompatible runtimes; individual teams build custom memory stores, bespoke tool interfaces, and ad hoc safety filters, resulting in duplicated effort and inconsistent governance practices. By open-sourcing the definition layer (AgenticFormat), agent specifications---termed ``Agent Cards''---can be shared, versioned, reviewed, and deployed on any compatible SDK or runtime, independent of the underlying model provider or infrastructure platform. The specification layer becomes a common artifact around which tooling, auditing processes, and interoperability standards can converge.

\subsection{Synergy with the Model Context Protocol}

The framework is designed for complementary operation with the \textbf{Model Context Protocol (MCP)}~\cite{mcp_anthropic}:

\begin{itemize}
    \item \textbf{MCP} standardizes \textit{how} an agent connects to external tools---defining the wire protocol, authentication flow, and data serialization format for tool connectors (e.g., connectors for Google Drive, Slack, or PostgreSQL).
    \item \textbf{AgenticFormat} standardizes \textit{who} the agent is---specifying its identity, permission scope, cognitive parameters, safety constraints, and the set of MCP servers it is authorized to access.
\end{itemize}

Decoupling agent identity and governance (AgenticFormat) from tool connectivity (MCP) enables independent evolution of both layers. New tool connectors can be developed and deployed without modifying agent specifications, and agent specifications can be updated---e.g., to tighten safety constraints or add new capabilities---without altering tool connector implementations.

\subsection{Cross-Language Portability}

Current agent frameworks (e.g., LangChain, AutoGen) are predominantly Python-based. Enterprise backend systems, however, frequently rely on Java for latency-sensitive services, where the JVM's static type system, mature concurrency primitives, and predictable garbage-collection behavior provide operational advantages.

Because the AgenticFormat Standard is language-agnostic, the Agentic AI Platform SDK can be implemented for any target language. The framework currently provides \textbf{\texttt{agentic-java}} alongside \textbf{\texttt{agentic-py}}---two SDK implementations that consume the same AgenticFormat blueprints. Java's type system---including Records for immutable data carriers, sealed interfaces for closed type hierarchies, and pattern matching for exhaustive case analysis---together with AgenticFormat schema validation, provides strong static guarantees at compile time. Agents specified in AgenticFormat can thus be deployed in low-latency, mission-critical Java microservices---e.g., real-time ad bidding, infrastructure autoscaling, transaction processing---rather than being limited to Python-based prototyping environments.

\section{Conclusion}
\label{sec:conclusion}

The transition from Generative AI to Agentic AI requires a corresponding shift in system architecture: from imperative scripts to \textbf{declarative definitions} that specify agent behavior as auditable data; from stateless, single-session interactions to \textbf{persistent cognitive architectures} that accumulate and consolidate experience over time; and from unconstrained stochastic output to \textbf{constraint manifolds} that enforce safety properties by construction rather than by post-hoc inspection.

The Auton Agentic AI Framework addresses each of these requirements through its constituent components. The \textbf{AgenticFormat Standard} provides a language-agnostic, declarative specification format that decouples agent identity from runtime execution. The formal agent model, grounded in an augmented POMDP with a latent reasoning space, provides a rigorous basis for analyzing agent behavior and enforcing the think-before-act execution discipline. The hierarchical memory architecture, with its Reflector-driven consolidation protocol, enables agents to retain and recall experience across sessions without unbounded context growth. The constraint manifold formalism ensures that safety properties are enforced through policy projection rather than brittle output filtering. The three-level self-evolution framework---spanning in-context adaptation, self-taught reasoning, and reinforcement learning---provides a path from static agent configurations to continuously improving autonomous systems. Runtime optimizations, including cognitive map-reduce parallelism, speculative execution, and attention-guided context pruning, address the latency constraints that otherwise limit deployment in interactive settings.

The framework is intended to support the deployment of reliable, auditable, and adaptive autonomous systems in enterprise environments where deterministic governance, cross-language portability, and measurable performance are operational requirements.


\begin{thebibliography}{99}

\bibitem{iti_agentic}
Information Technology Industry Council (ITI), ``Understanding Agentic AI.'' Accessed Jan 26, 2026. \url{https://www.itic.org/documents/artificial-intelligence/ITI_AgenticAI_Final.pdf}

\bibitem{mulani_medium}
Mulani, A., ``Building Agentic AI : Key Design Patterns,'' \textit{Medium}. \url{https://medium.com/@ashpaklmulani/agentic-ai-introduction-e91ad0ff7c06}

\bibitem{atla_frameworks}
``Comparing AI Agent Frameworks: A Guide to Building Reliable Agents,'' \textit{Atla AI}. \url{https://www.atla-ai.com/post/ai-agent-frameworks}

\bibitem{open_agent_spec}
Open Agent Specification Working Group, ``Open Agent Specification (Agent Spec) Technical Report,'' \textit{arXiv:2510.04173v3}. \url{https://arxiv.org/abs/2510.04173v3}

\bibitem{mcp_anthropic}
Anthropic, ``Code execution with MCP: building more efficient AI agents.'' Accessed Jan 26, 2026. \url{https://www.anthropic.com/engineering/code-execution-with-mcp}

\bibitem{mcp_explainer}
``What is MCP? (Model Context Protocol).'' Video Resource, Accessed Jan 26, 2026. \url{https://www.youtube.com/watch?v=pieK0dog66Q}

\bibitem{kaelbling1998pomdp}
Kaelbling, L.~P., Littman, M.~L., and Cassandra, A.~R., ``Planning and Acting in Partially Observable Stochastic Domains,'' \textit{Artificial Intelligence}, vol.~101, no.~1--2, pp.~99--134, 1998.

\bibitem{llm_guided_pomdp}
``LLM-Guided Probabilistic Program Induction for POMDP Model Estimation,'' \textit{arXiv:2505.02216v1}. \url{https://arxiv.org/html/2505.02216v1}

\bibitem{microsoft_reasoning}
Nambi, A., ``Unlocking Agentic Reasoning in LLMs,'' \textit{Microsoft Research}. \url{https://www.microsoft.com/en-us/research/people/akshayn/unlocking-agentic-reasoning-in-llms/}

\bibitem{yao2023react}
Yao, S., Zhao, J., Yu, D., Du, N., Shafran, I., Narasimhan, K., and Cao, Y., ``ReAct: Synergizing Reasoning and Acting in Language Models,'' in \textit{International Conference on Learning Representations (ICLR)}, 2023. \url{https://arxiv.org/abs/2210.03629}

\bibitem{wei2022cot}
Wei, J., Wang, X., Schuurmans, D., Bosma, M., Ichter, B., Xia, F., Chi, E., Le, Q.~V., and Zhou, D., ``Chain-of-Thought Prompting Elicits Reasoning in Large Language Models,'' in \textit{Advances in Neural Information Processing Systems (NeurIPS)}, 2022. \url{https://arxiv.org/abs/2201.11903}

\bibitem{princeton_consolidation}
``A model of autonomous interactions between hippocampus and neocortex driving sleep-dependent memory consolidation,'' \textit{Princeton Computational Memory Lab}. \url{https://compmem.princeton.edu/wp/wp-content/uploads/2022/10/A-model-of-autonomous-interactions-between-hippocampus-and-neocortex-driving-sleep-dependent-memory-consolidation.pdf}

\bibitem{memory_replay_bio}
``Memory replay in biological and artificial reinforcement learning,'' \textit{arXiv:2109.10034}. \url{https://arxiv.org/pdf/2109.10034}

\bibitem{aws_memory}
``Building smarter AI agents: AgentCore long-term memory deep dive,'' \textit{AWS Machine Learning Blog}. \url{https://aws.amazon.com/blogs/machine-learning/building-smarter-ai-agents-agentcore-long-term-memory-deep-dive/}

\bibitem{semantic_episodic_biorxiv}
``Semantic representations in episodic memory enhance recall and compositional consolidation,'' \textit{bioRxiv}. \url{https://www.biorxiv.org/content/10.1101/2025.10.03.680209v3.full.pdf}

\bibitem{memo_embodied}
``Memo: Training Memory-Efficient Embodied Agents with Reinforcement Learning,'' \textit{arXiv:2510.19732v1}. \url{https://arxiv.org/html/2510.19732v1}

\bibitem{safe_rl_manifold}
``Safe Reinforcement Learning on the Constraint Manifold: Theory and Applications,'' \textit{arXiv:2404.09080v1}. \url{https://arxiv.org/html/2404.09080v1}

\bibitem{formal_methods_verification}
``Formal Methods for Verification in Human-Agent Interaction,'' \textit{Diva-portal.org}. \url{http://www.diva-portal.org/smash/get/diva2:1950795/FULLTEXT01.pdf}

\bibitem{deny_monotone_access}
``Deny-monotone composition of hierarchical access control policies in distributed systems,'' \textit{ResearchGate}. \url{https://www.researchgate.net/publication/398484206}

\bibitem{survey_self_evolving}
``A Comprehensive Survey of Self-Evolving AI Agents: A New Paradigm Bridging Foundation Models and Lifelong Agentic Systems,'' \textit{arXiv:2508.07407}. \url{https://arxiv.org/abs/2508.07407}

\bibitem{sage_agent}
``SAGE: Self-evolving Agents with Reflective and Memory-augmented Abilities,'' \textit{arXiv:2409.00872v2}. \url{https://arxiv.org/html/2409.00872v2}

\bibitem{agentic_rl_demystified}
``Demystifying Reinforcement Learning in Agentic Reasoning,'' \textit{arXiv:2510.11701v1}. \url{https://arxiv.org/html/2510.11701v1}

\bibitem{lightman2023prm}
Lightman, H., Kosaraju, V., Burda, Y., Edwards, H., Baker, B., Lee, T., Leike, J., Schulman, J., Sutskever, I., and Cobbe, K., ``Let's Verify Step by Step,'' in \textit{International Conference on Learning Representations (ICLR)}, 2024. \url{https://arxiv.org/abs/2305.20050}

\bibitem{neurips_multi_turn}
``Reinforcing Multi-Turn Reasoning in LLM Agents via Turn-Level Reward Design and Credit Assignment,'' \textit{NeurIPS 2025}. \url{https://neurips.cc/virtual/2025/133311}

\bibitem{shinn2023reflexion}
Shinn, N., Cassano, F., Gopinath, A., Narasimhan, K., and Yao, S., ``Reflexion: Language Agents with Verbal Reinforcement Learning,'' in \textit{Advances in Neural Information Processing Systems (NeurIPS)}, 2023. \url{https://arxiv.org/abs/2303.11366}

\bibitem{langchain_reflection}
``Reflection Agents,'' \textit{LangChain Blog}. \url{https://www.blog.langchain.com/reflection-agents/}

\bibitem{star_reasoner}
Zelikman, E., et al., ``STaR: Self-Taught Reasoner,'' \textit{OpenReview}. Accessed Jan 26, 2026. \url{https://openreview.net/pdf?id=_3ELRdg2sgI}

\bibitem{cleaner_trajectory}
``CLEANER: Self-Purified Trajectories Boost Agentic Reinforcement Learning,'' \textit{arXiv:2601.15141}. \url{https://arxiv.org/html/2601.15141}

\bibitem{shao2024grpo}
Shao, Z., Wang, P., Zhu, Q., Xu, R., Song, J., Zhang, M., Li, Y.~K., Wu, Y., and Guo, D., ``DeepSeekMath: Pushing the Limits of Mathematical Reasoning in Open Language Models,'' \textit{arXiv:2402.03300}, 2024. \url{https://arxiv.org/abs/2402.03300}

\bibitem{guo2025deepseekr1}
Guo, D., Yang, D., Zhang, H., Song, J., Zhang, R., Xu, R., Zhu, Q., Ma, S., Wang, P., Bi, X., et al., ``DeepSeek-R1: Incentivizing Reasoning Capability in LLMs via Reinforcement Learning,'' \textit{arXiv:2501.12948}, 2025. \url{https://arxiv.org/abs/2501.12948}

\bibitem{schulman2017ppo}
Schulman, J., Wolski, F., Dhariwal, P., Radford, A., and Klimov, O., ``Proximal Policy Optimization Algorithms,'' \textit{arXiv:1707.06347}, 2017. \url{https://arxiv.org/abs/1707.06347}

\bibitem{k_level_policy}
``K-Level Policy Gradients for Multi-Agent Reinforcement Learning,'' \textit{arXiv:2509.12117v1}. \url{https://arxiv.org/html/2509.12117v1}

\end{thebibliography}
\end{document}